\documentclass[11pt,twoside]{article}
\usepackage{ap-article}
\usepackage{textcomp}
\def\labart{yourLabel}      
\shortauthors{P. Glauner \emph{et al.}}
\shorttitle{Challenge NTL Detection Survey}
\title{%
The Challenge of Non-Technical Loss Detection Using \\ Artificial Intelligence: A Survey
}
\author{%
Patrick Glauner\,\up{1},\
Jorge Augusto Meira\,\up{1},\
Petko Valtchev\,\up{12},\
Radu State\,\up{1},\
Franck Bettinger\,\up{3}
}
\addresses{%
\up{1}
Interdisciplinary Centre for Security, Reliability and Trust, University of Luxembourg,\\
4 Rue Alphonse Weicker,\\
Luxembourg, 2721, Luxembourg
\\ \vspace*{0.04truein}
E-mail: \{patrick.glauner, jorge.meira, radu.state\}@uni.lu
\\ \vspace*{0.05truein}
\up{2}
University of Quebec in Montreal,\\
PO Box 8888, Station Centre-ville,\\
Montreal, H3C 3P8, Canada\\ 
\vspace*{0.04truein}
E-mail: valtchev.petko@uqam.ca
\\ \vspace*{0.05truein}
\up{3}
CHOICE Technologies Holding S\`arl,\\
2-4 Rue Eugene Ruppert,\\
Luxembourg, 2453, Luxembourg \\ 
\vspace*{0.04truein}
E-mail: franck.bettinger@choiceholding.com
}
\begin{document}
\label{\labart-FirstPage}

\maketitle
\abstracts{%
Detection of non-technical losses (NTL) which include electricity theft, faulty meters or billing errors has attracted increasing attention from researchers in electrical engineering and computer science. NTLs cause significant harm to the economy, as in some countries they may range up to 40\% of the total electricity distributed.
The predominant research direction is employing artificial intelligence to predict whether a customer causes NTL. 
This paper first provides an overview of how NTLs are defined and their impact on economies, which include loss of revenue and profit of electricity providers and decrease of the stability and reliability of electrical power grids.
It then surveys the state-of-the-art research efforts in a up-to-date and comprehensive review of algorithms, features and data sets used.
It finally identifies the key scientific and engineering challenges in NTL detection and suggests how they could be addressed in the future.
}

\medskip

\vspace*{7pt}\textlineskip
\begin{multicols}{2}

\section{Introduction}
\label{sec:objectives}

Our modern society and daily activities strongly depend on the availability of electricity. Electrical power grids allow to distribute and deliver electricity from generation infrastructure such as power plants or solar cells to customers such as residences or factories.
One frequently appearing problem are losses in power grids, which can be classified into two categories: technical and non-technical losses.

Technical losses occur mostly due to power dissipation. This is naturally caused by internal electrical resistance and the affected components include generators, transformers and transmission lines.

The complementary non-technical losses (NTL) are primarily caused by electricity theft. In most countries, NTLs account for the predominant part of the overall losses as discussed in Ref. \citen{nagi2010nontechnical}. Therefore, it is most beneficial to first reduce NTLs before reducing technical losses as proposed in Ref. \citen{amin2012incentives}. In particular, NTLs include, but are not limited to, the following causes reported in Refs. \citen{chauhan2013non} and \citen{smith2004electricity}:
\begin{itemize}
\item Meter tampering in order to record lower consumptions
\item Bypassing meters by rigging lines from the power source
\item Arranged false meter readings by bribing meter readers
\item Faulty or broken meters
\item Un-metered supply
\item Technical and human errors in meter readings, data processing and billing
\end{itemize}

NTLs cause significant harm to economies, including loss of revenue and profit of electricity providers, decrease of the stability and reliability of electrical power grids and extra use of limited natural resources which in turn increases pollution.
For example, in India, NTLs are estimated at US\$ 4.5 billion in Ref. \citen{bhatia2004reforming}. NTLs are simultaneously reported in Refs. \citen{depuru2013high} and \citen{nagi2011improving} to range up to 40\% of the total electricity distributed in countries such as Brazil, India, Malaysia or Lebanon. They are also of relevance in developed countries, for example estimates of NTLs in the UK and US that range from US\$ 1-6 billion are reported in Refs. \citen{nagi2010nontechnical} and \citen{alam2004power}.

We want to highlight that only few works on NTL detection have been reported in the literature in the last three to four years. Given that NTL detection is an active field in industrial R\&D, it is to our surprise that academic research in this field has dropped in the last few years.

From an electrical engineering perspective, one method to detect losses is to calculate the energy balance reported in Ref. \citen{oliveira2001new}, which requires topological information of the network. In emerging economies, which are of particular interest due to their high NTL proportion, this is not realistic for the following reasons: (i) network topology undergoes continuous changes in order to satisfy the rapidly growing demand of electricity, (ii) infrastructure may break and lead to wrong energy balance calculations and (iii) it requires transformers, feeders and connected meters to be read at the same time.

A more flexible and adaptable approach is to employ artificial intelligence (AI), which is well covered in Ref. \citen{russell2009artificial}. AI allows to analyze customer profiles, their data and known irregular behavior. This allows to trigger possible inspections of customers that have abnormal electricity consumption patterns.
Technicians then carry out inspections, which allow them to remove possible manipulations or malfunctions of the power infrastructure. Furthermore, the fraudulent customers can be charged for the additional electricity consumed.
However, carrying out inspections is costly, as it requires physical presence of technicians. 

NTL detection methods reported in the literature fall into two categories: expert systems and machine learning.
Expert systems incorporate hand-crafted rules for decision making. In contrast, machine learning gives computers the ability to learn from examples without being explicitly programmed.
Historically, NTL detection systems were based on domain-specific rules. However, over the years, the field of machine learning has become the predominant research direction of NTL detection.
To date, there is no authoritative survey that compares the various approaches of NTL detection methods reported in the literature. We are also not aware of any existing survey that discusses the shortcomings of the state of the art.
In order to advance in NTL detection, the main contributions of this survey are the following:
\begin{itemize}
\item We provide a detailed review and critique of state-of-the-art NTL detection research employing AI methods in Section~\ref{chapter:stateoftheart}.
\item We identify the unsolved key challenges of this field in Section~\ref{chapter:objectives}.
\item We describe in detail the proposed methods to solve the most relevant challenges in the future in Section~\ref{chapter:method}.
\item We put these challenges in the context of AI research as a whole as they are of relevance to many other learning and anomaly detection problems.
\end{itemize}

\section{The State of the Art}
\label{chapter:stateoftheart}

NTL detection can be treated as a special case of fraud detection, for which general surveys are provided in Refs. \citen{bolton2002statistical} and \citen{kou2004survey}. Both highlight expert systems and machine learning as key methods to detect fraudulent behavior in applications such as credit card fraud, computer intrusion and telecommunications fraud.
This section is focused on an overview of the existing AI methods for detecting NTLs.
Existing short surveys of the past efforts in this field, such as in Refs. \citen{chauhan2013non,kazerooni2014literature,jiang2014energy} and \citen{mclaughlin2009energy} only provide a narrow comparison of the entire range of relevant publications. The novelty of this survey is to not only review and compare a wide range of results reported in the literature, but to also derive the unsolved challenges of NTL detection.

\subsection{Features}
In this subsection, we summarize and group the features reported in the literature.

\subsubsection{Monthly consumption}
Many works on NTL detection use traditional meters, which are read monthly or annually by meter readers. Based on this data, average consumption features are used in Refs. \citen{nagi2010nontechnical, nagi2011improving, glauner2016large, nagi2008non} and \citen{nagi2008detection}. In those works, the feature computation used can be summarized as follows: For $M$ customers $\{0, 1, ..., M - 1\}$ over the last $N$ months $\{0, 1, ..., N -1\}$, a feature matrix $F$ is computed, in which element $F_{m, d}$ is a daily average kWh consumption feature during that month:
\begin{align}
x_d^{(m)} = \frac{L_d^{(m)}}{R^{(m)}_{d} - R^{(m)}_{d-1}},
\end{align}
where for customer $m$, $L_d^{(m)}$ is the kWh consumption increase between the meter reading to date $R^{(m)}_d$ and the previous one $R^{(m)}_{d-1}$. $R^{(m)}_{d} - R^{(m)}_{d-1}$ is the number of days between both meter readings of customer $m$.

The previous 24 monthly meter readings are used in Refs. \citen{muniz2009neuro} and \citen{muniz2009irregularity}. The features computed are the monthly consumption before the inspection, the consumption in the same month in the year before the consumption in the past three months and the customer's consumption over the past 24 months.
Using the previous six monthly meter readings, the following features are derived in Ref. \citen{angelos2011detection}: average consumption, maximum consumption, standard deviation, number of inspections and the average consumption of the residential area.
The average consumption is also used as a feature in Refs. \citen{costa2013fraud} and \citen{spiric2014using}.

\subsubsection{Smart meter consumption}
With the increasing availability of smart meter devices, consumption of electric energy in short intervals can be recorded. Consumption features of intervals of 15 minutes are used in Refs. \citen{cabral2009fraud} and \citen{depuru2011support}, whereas intervals of 30 minutes are used in Refs. \citen{nagi2010ntl} and \citen{sahoo2015electricity}.

The $4\times 24=96$ measurements of Ref. \citen{depuru2011support} are encoded to a 32-dimensional space in Refs. \citen{depuru2013high} and \citen{depuru2012enhanced}. Each measurement is 0 or positive. Next, it is then mapped to 0 or 1, respectively. Last, the 32 features are computed. A feature is the weighted sum of three subsequent values, in which the first value is multiplied by 4, the second by 2 and the third by 1.

The maximum consumption in any 15-minute period is used as a feature in Refs. \citen{ramos2012identification, ramos2009fast, ramos2011new} and \citen{ramos2010learning}. The load factor is computed by dividing the demand contracted by the maximum consumption.

Features from the consumption time series called shape factors are derived from the consumption time series including the impact of lunch times, nights and weekends in Ref. \citen{nizar2006customer}.

\subsubsection{Master data}
Master data represents customer reference data such as name or address, which typically changes infrequently.
The work in Ref. \citen{costa2013fraud} uses the following features from the master data for classification: location (city and neighborhood), business class (e.g. residential or industrial), activity type (e.g. residence or drugstore), voltage (110V or 200V), number of phases (1, 2 or 3) and meter type.
The demand contracted, which is the number of kW of continuous availability requested from the energy company and the total demand in kW of installed equipment of the customer are used in Refs. \citen{ramos2010learning, ramos2009fast, ramos2011new}.
In addition, information about the power transformer to which the customer is connected to is used in Ref. \citen{ramos2012identification}.
The town or customer in which the customer is located, the type of voltage (low, median or high), the electricity tariff, the contracted power as well as the number of phases (1 or 3) are used in Ref. \citen{spiric2014using}.
Related master data features are used in Ref. \citen{nizar2006customer}, including the type of customer, location, voltage level, type of climate (rainy or hot), weather conditions and type of day.

\subsubsection{Credit worthiness}
The works in Refs. \citen{nagi2010nontechnical, nagi2008non} and \citen{nagi2008detection} use the credit worthiness ranking (CWR) of each customer as a feature. It is computed from the electricity provider's billing system and reflects if a customer delays or avoids payments of bills. CWR ranges from 0 to 5 where 5 represents the maximum score. It reflects different information about a customer such as payment performance, income and prosperity of the neighborhood in a single feature.

\subsection{Expert systems and fuzzy systems}
An ensemble pre-filters the customers to select irregular and regular customers in Ref. \citen{muniz2009neuro}. These customers are then used for training as they represent well the two different classes. This is done because of noise in the inspection labels. In the classification step, a neuro-fuzzy hierarchical system is used. In this setting, a neural network is used to optimize the fuzzy membership parameters, which is a different approach to the stochastic gradient descent method used in Ref. \citen{glauner2016large}. A precision of 0.512 and an accuracy of 0.682 on the test set are obtained.

Profiles of 80K low-voltage and 6K high-voltage customers in Malaysia having meter readings every 30 minutes over a period of 30 days are used in Ref. \citen{nagi2010ntl} for electricity theft and abnormality detection. A test recall of 0.55 is reported.
This work is related to features of Ref. \citen{nagi2011improving}, however, it uses entirely fuzzy logic incorporating human expert knowledge for detection.

The work in Ref. \citen{nagi2010nontechnical} is combined with a fuzzy logic expert system postprocessing the output of the SVM in Ref. \citen{nagi2011improving} for \texttildelow 100K customers. The motivation of that work is to integrate human expert knowledge into the decision making process in order to identify fraudulent behavior. A test recall of 0.72 is reported.

Five features of customers' consumption of the previous six months are derived in Ref. \citen{angelos2011detection}: average consumption, maximum consumption, standard deviation, number of inspections and the average consumption of the residential area. These features are then used in a fuzzy c-means clustering algorithm to group the customers into c classes. Subsequently, the fuzzy membership values are used to classify customers into NTL and non-NTL using the Euclidean distance measure. On the test set, an average precision (called average assertiveness) of 0.745 is reported.

\subsection{Neural networks}
Neural networks are loosely inspired by how the human brain works and allow to learn complex hypotheses from data. They are well described for example in Ref. \citen{bishop1996neural}.
Extreme learning machines (ELM) are one-hidden layer neural networks in which the weights from the inputs to the hidden layer are randomly set and never updated. Only the weights from the hidden to output layer are learned. The ELM algorithm is applied to NTL detection in meter readings of 30 minutes in Ref. \citen{nizar2008power}, for which a test accuracy of 0.5461 is reported.

An ensemble of five neural networks (NN) is trained on samples of a data set containing \texttildelow 20K customers in Ref. \citen{muniz2009irregularity}. Each neural network uses features calculated from the consumption time series plus customer-specific pre-computed attributes. A precision of 0.626 and an accuracy of 0.686 are obtained on the test set.

A self-organizing map (SOM) is a type of unsupervised neural network training algorithm that is used for clustering. SOMs are applied to weekly customer data of 2K customers consisting of meter readings of 15 minutes in Ref. \citen{cabral2009fraud}. This allows to cluster customers' behavior into fraud or non-fraud. Inspections are only carried out if certain hand-crafted criteria are satisfied including how well a week fits into a cluster and if no contractual changes of the customer have taken place. A test accuracy of 0.9267, a test precision of 0.8526, and test recall of 0.9779 are reported.

A data set of \texttildelow 22K customers is used in Ref. \citen{costa2013fraud} for training a neural network. It uses the average consumption of the previous 12 months and other customer features such as location, type of customer, voltage and whether there are meter reading notes during that period. On the test set, an accuracy of 0.8717, a precision of 0.6503 and a recall of 0.2947 are reported.

\subsection{Support vector machines}
The Support Vector Machines (SVM) introduced in Ref. \citen{vapnik1999overview} is a state-of-the-art classification algorithm that is less prone to overfitting.
Electricity customer consumption data of less than 400 highly imbalanced out of \texttildelow 260K customers in Kuala Lumpur, Malaysia are used in Ref. \citen{nagi2008non}. Each customer has 25 monthly meter readings in the period from June 2006 to June 2008. From these meter readings, daily average consumption features per month are computed. Those features are then normalized and used for training in a SVM with a Gaussian kernel. In addition, credit worthiness ranking (CWR) is used. It is computed from the electricity provider's billing system and reflects if a customer delays or avoids payments of bills. CWR ranges from 0 to 5 where 5 represents the maximum score. It was observed that CWR is a significant indicator of whether customers commit electricity theft.
For this setting, a recall of 0.53 is achieved on the test set.
A related setting is used in Ref. \citen{nagi2010nontechnical}, where a test accuracy of 0.86 and a test recall of 0.77 are reported on a different data set.

SVMs are also applied to 1,350 Indian customer profiles in Ref. \citen{depuru2011support}. They are split into 135 different daily average consumption patterns, each having 10 customers. For each customer, meters are read every 15 minutes. A test accuracy of 0.984 is reported. This work is extended in Ref. \citen{depuru2012enhanced} by encoding the $4\times 24=96$-dimensional input in a lower dimension indicating possible irregularities. This encoding technique results in a simpler model that is faster to train while not losing the expressiveness of the data and results in a test accuracy of 0.92.

Consumption profiles of 5K Brazilian industrial customer profiles are analyzed in Ref. \citen{ramos2012identification}. Each customer profile contains 10 features including the demand billed, maximum demand, installed power, etc. In this setting, a SVM slightly outperforms K-nearest neighbors (KNN) and a neural network, for which test accuracies of 0.9628, 0.9620 and 0.9448, respectively, are reported.

The work of Ref. \citen{depuru2012enhanced} is extended in Ref. \citen{depuru2013high} by introducing high performance computing algorithms in order to enhance the performance of the previously developed algorithms. This faster model has a test accuracy of 0.89.

A data set of \texttildelow 700K Brazilian customers, \texttildelow 31M monthly meter readings from January 2011 to January 2015 and \texttildelow 400K inspection data is used in Ref. \citen{glauner2016large}. It employs an industrial Boolean expert system, its fuzzified extension and optimizes the fuzzy system parameters using stochastic gradient descent described in Ref. \citen{bottou2004stochastic} to that data set. This fuzzy system outperforms the Boolean system. Inspired by Ref. \citen{nagi2008non}, a SVM using daily average consumption features of the last 12 months performs better than the expert systems, too. The three algorithms are compared to each other on samples of varying fraud proportion containing \texttildelow 100K customers. It uses the area under the (receiver operating characteristic) curve (AUC), which is discussed in Section~\ref{chapter:challenges:imbalance}. 
For a NTL proportion of 5\%, it reports AUC test scores of 0.465, 0.55 and 0.55 for the Boolean system, optimized fuzzy system and SVM, respectively. 
For a NTL proportion of 20\%, it reports AUC test scores of 0.475, 0.545 and 0.55 for the Boolean system, optimized fuzzy system and SVM, respectively.

\subsection{Genetic algorithms}
The work in Refs. \citen{nagi2010nontechnical} and \citen{nagi2008non} is extended by using a genetic SVM for 1,171 customers in Ref. \citen{nagi2008detection}. It uses a genetic algorithm in order to globally optimize the hyperparameters of the SVM. Each chromosome contains the Lagrangian multipliers $(\alpha_1, ..., \alpha_i)$, regularization factor $C$ and Gaussian kernel parameter $\gamma$. This model achieves a test recall of 0.62.
 
A data set of \texttildelow 1.1M customers is used in Ref. \citen{costa2013optimization}. The paper identifies the much smaller class of inspected customers as the main challenge in NTL detection. It then proposes stratified sampling in order to increase the number of inspections and to minimize the statistical variance between them. The stratified sampling procedure is defined as a non-linear restricted optimization problem of minimizing the overall energy loss due to electricity theft. This minimization problem is solved using two methods: (1) genetic algorithm and (2) simulated annealing. The first approach outperforms the other one. Only the reduced variance is reported, which is not comparable to the other research and therefore left out of this survey.

\subsection{Rough sets}
Rough sets give lower and upper approximations of an original conventional or crisp set. The first application of rough set analysis applied to NTL detection is described in Ref. \citen{cabral2004rough} on 40K customers, but lacks details on the attributes used per customer, for which a test accuracy of 0.2 is achieved.
Rough set analysis is also applied to NTL detection in Ref. \citen{spiric2014using} on features related to Ref. \citen{costa2013fraud}. This supervised learning technique allows to approximate concepts that describe fraud and regular use. A test accuracy of 0.9322 is reported.

\subsection{Other methods}
Different feature selection techniques for customer master data and consumption data are assessed in Ref. \citen{nizar2006customer}. Those methods include complete search, best-first search, genetic search and greedy search algorithms for the master data. Other features called shape factors are derived from the consumption time series including the impact of lunch times, nights and weekends on the consumption. These features are used in K-means for clustering similar consumption time series. In the classification step, a decision tree is used to predict whether a customer causes NTLs or not. An overall test accuracy of 0.9997 is reported.

Optimum path forests (OPF), a graph-based classifier, is used in Ref. \citen{ramos2011new}. It builds a graph in the feature space and uses so-called ``prototypes" or training samples. Those become roots of their optimum-path tree node. Each graph node is classified based on its most strongly connected prototype. This approach is fundamentally different to most other learning algorithms such as SVMs or neural networks which learn hyperplanes. Optimum path forests do not learn parameters, thus making training faster, but predicting slower compared to parametric methods. They are used in Ref. \citen{ramos2009fast} for 736 customers and achieved a test accuracy of 0.9021, outperforming SMVs with Gaussian and linear kernels and a neural network which achieved test accuracies of 0.8893, 0.4540 and 0.5301, respectively. Related results and differences between these classifiers are also reported in Ref. \citen{ramos2010learning}.

A different method is to estimate NTLs by subtracting an estimate of the technical losses from the overall losses reported in Ref. \citen{sahoo2015electricity}. It models the resistance of the infrastructure in a temperature-dependent model using regression which approximates the technical losses. It applies the model to a data set of 30 customers for which the consumption was recorded for six days with meter readings every 30 minutes for theft levels of 1, 2, 3, 4, 6, 8 and 10\%. The respective test recalls in linear circuits are 0.2211, 0.7789, 0.9789, 1, 1, 1 and 1, respectively.

\subsection{Summary}
A summary and comparison of models, data sets and performance measures of selected work discussed in this review is reported in Table~\ref{table:summary}.
The most commonly used models comprise Boolean and fuzzy expert systems, SVMs and neural networks. In addition, genetic methods, OPF and regression methods are used.
Data set sizes have a wide range from 30 up to 700K customers. However, the largest data set of 1.1M customers in Ref. \citen{costa2013optimization} is not included in the table because only the variance is reduced and no other performance measure is provided.
Accuracy and recall are the most popular performance measures in the literature, ranging from 0.45 to 0.99 and from 0.29 to 1, respectively. Only very few publications report the precision of their models, ranging from 0.51 to 0.85. The AUC is only reported in one publication.
The challenges of finding representative performance measures and how to compare individual contributions are discussed in Sections~\ref{chapter:challenges:imbalance} and \ref{chapter:challenges:comparison}, respectively.

\begin{table*}\centering 
\tcap{Summary of models, data sets and performance measures (two-decimal precision).}
{\small
\setlength{\extrarowheight}{3pt}
\baselineskip=13pt
\begin{tabular}{l@{\qquad}l@{\qquad}l@{\qquad}l@{\qquad}l@{\qquad}l@{\qquad}l@{\qquad}l}\\
\hline
Ref. & Model & \#Customers & Accuracy & Precision & Recall & AUC & NTL/theft proportion \\
\hline
\citen{nagi2010nontechnical} & SVM (Gauss) & $< 400$ & 0.86 & - & 0.77 & - & - \\
\citen{nagi2011improving} & SVM + fuzzy & 100K & - & - & 0.72 & - & - \\
\citen{glauner2016large} & Bool rules & 700K & - & - & - & 0.47 & 5\% \\
\citen{glauner2016large} & Fuzzy rules & 700K & - & - & - & 0.55 & 5\% \\
\citen{glauner2016large} & SVM (linear) & 700K & - & - & - & 0.55& 5\% \\
\citen{glauner2016large} & Bool rules & 700K & - & - & - & 0.48 & 20\% \\
\citen{glauner2016large} & Fuzzy rules & 700K & - & - & - & 0.55 & 20\% \\
\citen{glauner2016large} & SVM (linear) & 700K & - & - & - & 0.55& 20\% \\
\citen{nagi2008non} & SVM & $< 400$ & - & - & 0.53 & - & - \\
\citen{nagi2008detection} & Genetic SVM & 1,171 & - & - & 0.62 & - & - \\
\citen{muniz2009neuro} & Neuro-fuzzy & 20K & 0.68 & 0.51 & - & - & - \\
\citen{costa2013fraud} & NN & 22K & 0.87 & 0.65 & 0.29 & - & - \\
\citen{spiric2014using} & Rough sets & N/A & 0.93 & - & - & - & - \\
\citen{cabral2009fraud} & SOM & 2K & 0.93 & 0.85 & 0.98 & - &  - \\
\citen{depuru2011support} & SVM (Gauss) & 1,350 & 0.98 & - & - & - & - \\
\citen{sahoo2015electricity} & Regression & 30 & - & - & 0.22 & - & 1\% \\
\citen{sahoo2015electricity} & Regression & 30 & - & - & 0.78 & - & 2\% \\
\citen{sahoo2015electricity} & Regression & 30 & - & - & 0.98 & - & 3\% \\
\citen{sahoo2015electricity} & Regression & 30 & - & - & 1 & - & 4-10\% \\
\citen{ramos2012identification} & SVM & 5K & 0.96 & - & - & - & - \\
\citen{ramos2012identification} & KNN & 5K & 0.96 & - & - & - & - \\
\citen{ramos2012identification} & NN & 5K & 0.94 & - & - & - & - \\
\citen{ramos2009fast} & OPF & 736 & 0.90 & - & - & - & - \\
\citen{ramos2009fast} & SVM (Gauss) & 736 & 0.89 & - & - & - & - \\
\citen{ramos2009fast} & SVM (linear) & 736 & 0.45 & - & - & - & - \\
\citen{ramos2009fast} & NN & 736 & 0.53 & - & - & - & - \\
\citen{nizar2006customer} & Decision tree & N/A & 0.99 & - & - & - & - \\
\hline
\end{tabular}}\label{table:summary}
\end{table*}

\section{Challenges}
\label{chapter:objectives}

The research reviewed in the previous section indicates multiple open challenges. These challenges do not apply to single contributions, rather they spread across different ones.
In this section, we discuss these challenges, which must be addressed in order to advance in NTL detection. Concretely, we discuss common topics that have not yet received the necessary attention in previous research and put them in the context of AI research as a whole.

\subsection{Class imbalance and evaluation metric}
\label{chapter:challenges:imbalance}
Imbalanced classes appear frequently in machine learning, which also affects the choice of evaluation metrics as discussed in Refs. \citen{japkowicz2002class} and \citen{tang2009svms}.
Most NTL detection research do not address this property. Therefore, in many cases, high accuracies or high recalls are reported, such as in Refs. \citen{nagi2008non,costa2013fraud,spiric2014using,ramos2011new} and \citen{costa2013optimization}.
The following examples demonstrate why those performance measures are not suitable for NTL detection in imbalanced data sets: for a test set containing 1K customers of which 999 have regular use, (1) a classifier always predicting non-NTL has an accuracy of 99.9\%, whereas  in contrast, (2) a classifier always predicting NTL has a recall of 100\%.
While the classifier of the first example has a very high accuracy and intuitively seems to perform very well, it will never predict any NTL. In contrast, the classifier of the second example will find all NTL, but triggers many costly and unnecessary physical inspections by inspecting all customers.
This topic is addressed rarely in NTL literature, such as in Refs. \citen{muniz2009irregularity} and \citen{di2012improving}, and these contributions do not use a proper single measure of performance of a classifier when applied to an imbalanced data set.

\subsection{Feature description}
Generally, hand-crafting features from raw data is a long-standing issue in machine learning having significant impact on the performance of a classifier, as discussed in Ref. \citen{domingos2012few}.
Different feature description methods have been reviewed in the previous section. They fall into two main categories: features computed from the consumption profile of customers, which are from monthly meter readings, for example in Refs. \citen{nagi2010nontechnical,nagi2011improving,glauner2016large,nagi2008non,nagi2008detection,muniz2009neuro,muniz2009irregularity,angelos2011detection,costa2013fraud} and \citen{spiric2014using}, or smart meter readings, for example in Refs. \citen{depuru2013high,cabral2009fraud,depuru2011support,nagi2010ntl,sahoo2015electricity,depuru2012enhanced,ramos2012identification,ramos2009fast,ramos2011new,ramos2010learning}, and \citen{nizar2006customer}, and features from the customer master data in Refs. \citen{costa2013fraud,spiric2014using,ramos2012identification,ramos2009fast,ramos2011new,ramos2010learning} and \citen{nizar2006customer}. 
The features computed from the time series are very different for monthly meter readings and smart meter readings.  The results of those works are not easily interchangeable. While electricity providers continuously upgrade their infrastructure to smart metering, there will be many remaining traditional meters. In particular, this applies to emerging countries.

There are only few works on assessing the statistical usefulness of features for NTL detection, such as in Ref. \citen{glauner2016neighborhood}. 
Almost all works on NTL detection define features and subsequently report improved models that were mostly found  experimentally without having a strong theoretical foundation.

\subsection{Data quality}
In the preliminary work of Ref. \citen{glauner2016large}, we noticed that the inspection result labels in the training set are not always correct and that some fraudsters may be labelled as non-fraudulent. The reasons for this may include bribing, blackmailing or threatening of the technician performing the inspection. Also, the fraud may be done too well and is therefore not observable by technicians. Another reason may be incorrect processing of the data. It must be noted that the latter reason may, however, also label non-fraudulent behavior as fraudulent.
Handling noise is a common challenge in machine learning. In supervised machine learning settings, most existing methods address handling noise in the input data. There are different regularization methods such as $L_1$ or $L_2$ regularization discussed in Ref. \citen{ng2004feature} or learning of invariances allowing learning algorithms to better handle noise in the input data discussed in Refs. \citen{bishop2006pattern} and \citen{le1990handwritten}. However, handling noise in the training labels is less commonly addressed in the machine learning literature.
Most NTL detection research use supervised methods. This shortcoming of the training data and potential wrong labels in particular are only rarely reported in the literature, such as in Ref. \citen{muniz2009neuro}, which uses an ensemble to pre-filter the training data.

\subsection{Covariate shift}
Covariate shift refers to the problem of training data (i.e. the set of inspection results) and production data (i.e. the set of customers to generate inspections for) having different distributions. This fact leads to unreliable NTL predictors when learning from this training data.
Historically, covariate shift has been a long-standing issue in statistics, as surveyed in Ref. \citen{harford2014big}. For example, The Literary Digest sent out 10M questionnaires in order to predict the outcome of the 1936 US Presidential election. They received 2.4M returns. Nonetheless, the predicted result proved to be wrong. The reason for this was that they used car registrations and phone directories to compile a list of recipients. In that time, the households that had a phone or a car represented a biased sample of the overall population. In contrast, George Gallup only interviewed 3K handpicked people, which were an unbiased sample of the population. As a consequence, Gallup could predict the outcome of the election very well.

For about the last fifteen years, the Big Data paradigm followed in machine learning has been to gather more data rather than improving models. Hence, one may assume that having simply more customer and inspection data would help to detect NTL more accurately. However, in many cases, the data may be biased as depicted in Fig.~\ref{fig:example}. 

\begin{Figure}
\includegraphics[width=.8\columnwidth]{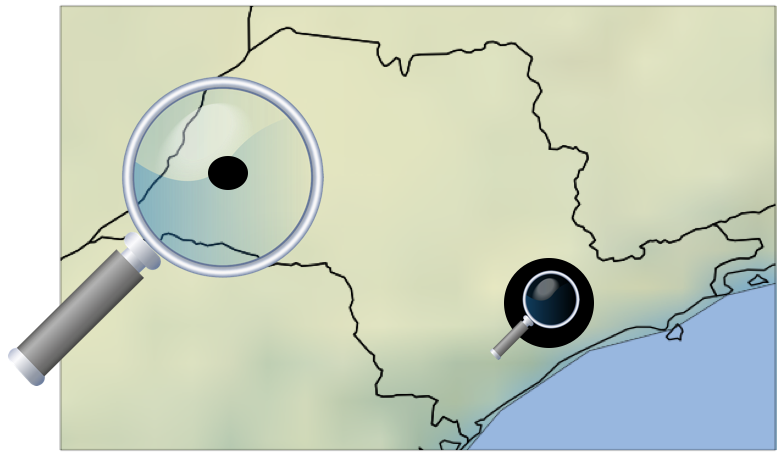} 
\fcaption{Example of spatial bias: The large city is close to the sea, whereas the small city is located in the interior of the country. The weather in the small city undergoes stronger changes during the year. The subsequent change of electricity consumption during the year triggers many inspections. As a consequence, most inspections are carried out in the small city. Therefore, the sample of customers inspected does not represent the overall population of customers.}
\label{fig:example}
\end{Figure}

One reason is, for example, that electricity suppliers previously focused on certain neighborhoods for inspections.
Concretely, the customers inspected are a sample of the overall population of customers. In this example, there is a spatial bias. Hence, the inspections do not represent the overall population of customers. As a consequence, when learning from the inspection results, a bias is learned, making predictions less reliable.
Aside from spatial covariate shift, there may be other types of covariate shift in the data, such as the meter type, connection type, etc.

To the best of our knowledge, the issue of covariate change has not been addressed in the literature on NTL detection. However, in many cases it may lead to unreliable NTL detection models.
Therefore, we consider it important to derive methods for quantifying and reducing the covariate shift in data sets relevant to NTL detection. This will allow to build more reliable NTL detection models.

\subsection{Scalability}
The number of customers used throughout the research reviewed significantly varies.
For example, Refs. \citen{nagi2008non} and \citen{sahoo2015electricity} only use less than a few hundred customers in the training. A SVM with a Gaussian kernel is used in Ref. \citen{nagi2008non}. In that setting, training is only feasible in a realistic amount of time for up to a couple of tens of thousands of customers in current implementations as discussed in Ref. \citen{CC01a}.
A regression model using the Moore-Penrose pseudoinverse introduced in Ref. \citen{penrose1955generalized} is used in Ref. \citen{sahoo2015electricity}. This model is also only able to scale to up to a couple of tens of thousands of customers.
Neural networks are trained on up to a couple of tens of thousands of customers in Refs. \citen{muniz2009irregularity} and \citen{costa2013fraud}. The training methods used in prior work usually do not scale to significantly larger customer data sets. Larger data sets using up to hundreds of thousands or millions of customers are used in Refs. \citen{glauner2016large} and \citen{costa2013optimization} using a SVM with linear kernel or genetic algorithms, respectively.
An important property of NTL detection methods is that their computational time must scale to large data sets of hundreds of thousands or millions of customers. Most works reported in the literature do not satisfy this requirement.

\subsection{Comparison of different methods}
\label{chapter:challenges:comparison}
Comparing the different methods reviewed in this paper is challenging because they are tested on different data sets, as summarized in Table~\ref{table:summary}. In many cases, the description of the data lacks fundamental properties such as the number of meter readings per customer, NTL proportion, etc. In order to increase the reliability of a comparison, joint efforts of different research groups are necessary. These efforts need to address the benchmarking and comparability of NTL detection systems based on a comprehensive freely available data set.

\section{Suggested Methodology}
\label{chapter:method}
We have reviewed state-of-the-art research in machine learning and identified the following suggested methodology for solving the main research challenges in NTL detection:

\subsection{Handling class imbalance and evaluation metric}
How can we handle the imbalance of classes and assess the outcome of classifications using accurate metrics?

Anomaly detection problems are particularly imbalanced, meaning that there are much more training examples of the regular class compared to the anomaly class.
Most works on NTL detection do not reflect the imbalance and simply report accuracies or recalls, for example in Refs. \citen{nagi2008non,costa2013fraud,spiric2014using,ramos2011new} and \citen{costa2013optimization}.
This is also depicted in Table~\ref{table:summary}.
For NTL detection, the goal is to reduce the false positive rate (FPR) to decrease the number of costly inspections, while increasing the true positive rate (TPR) to find as many NTL occurrences as possible. In Ref. \citen{glauner2016large}, we propose to use a receiver operating characteristic (ROC) curve, which plots the TPR against the FPR. The area under the curve (AUC) is a performance measure between $0$ and $1$, where any binary classifier with an AUC $> 0.5$ performs better than random guessing. 
In order to assess a NTL prediction model using a single performance measure, the AUC was picked as the most suitable one in Ref. \citen{glauner2016large}.

All works in the literature only use a fixed NTL proportion in the data set, for example in Refs.  \citen{nagi2008non,muniz2009irregularity,costa2013fraud,spiric2014using,ramos2011new,costa2013optimization} and \citen{di2012improving}.
We think that it is necessary to investigate more into this topic in order to report reliable and imbalance-independent results that are valid for different levels of imbalance. 
This will allow to build models that work in different regions, such as in regions with a high NTL ratio as well as in regions with a low occurrence of NTLs. Therefore, we suggest to create samples of different NTL proportions and assess the models on the entire range of these samples.
In the preliminary work of Ref. \citen{glauner2016large}, we also noticed that the precision usually grows linearly with the NTL proportion in the data set. It is therefore not suitable for low NTL proportions. However, we did not notice this for the recall and made observations of non-linearity similar to related work in Ref. \citen{sahoo2015electricity}, as depicted in Table~\ref{table:summary}.
With the limitations of precision and recall, the $F_1$ score did not prove to work as a reliable performance measure.

Furthermore, we suggest to derive multi-criteria evaluation metrics for NTL detection and rank customers that cause a NTL with a confidence level, for example models related to the ones in introduced in Ref. \citen{lopez2016protocol}. For example, the criteria we suggest to include are the costs of inspections and possible increases in revenue.

\subsection{Feature description and modeling temporal behavior}
How can we describe features that accurately reflect NTL occurrence and can we self-learn these features from data? NTL of customers is a set of inherently temporal events where for example a fraud of customers excites themselves or other related customers to commit fraud as well. How can we extend temporal processes to model the characteristics of NTL?

Most research on NTL uses primarily information from the consumption time series. The consumption is from traditional meters, such as in Refs. \citen{nagi2010nontechnical,glauner2016large,nagi2008non,muniz2009neuro} and \citen{muniz2009irregularity}, or smart meters, such as in Refs. \citen{depuru2013high,depuru2011support,nagi2010ntl,sahoo2015electricity,ramos2009fast,ramos2010learning} and \citen{nizar2006customer}.
Both meter types will co-exist in the next decade and the results of those works are not easily interchangeable. Therefore, we suggest to shift to self-learning of features from the consumption time series. This topic has not been explored in the literature on NTL detection yet.
Deep learning allows to self-learn hidden correlations and increasingly more complex feature hierarchies from the raw data input as discussed in Ref. \citen{lecun2015deep}. This approach has lead to breakthroughs in image analysis and speech recognition as presented in Ref. \citen{hinton2012deep}. One possible method to overcome the challenge of feature description for NTL detection is by finding a way to apply deep learning to it.

In a different vein, we believe that the neighborhood of customers contains information about whether a customer may cause a NTL or not. Our hypothesis is confirmed by initial work described in Ref. \citen{angelos2011detection}, in which also the average consumption of the residential neighborhood is used for classification of NTL. We have shown in Ref. \citen{glauner2016neighborhood} that features derived from the inspection ratio and NTL ratio in a neighborhood help to detect NTL.

A temporal process, such as a Hawkes process described in Ref. \citen{laub2015hawkes}, models the occurrence of an event that depends on previous events. Hawkes processes include self-excitement, meaning that once an event happens, that event is more likely to happen in the near future again and decays over time. In other words, the further back the event in the process, the less impact it has on future events.
The dynamics of Hawkes processes look promising for modeling NTL: Our first hypothesis is that once customers were found to steal electricity, finding them or their neighbors to commit theft again is more likely in the near future again and decays over time. A Hawkes process allows to model this first hypothesis.
Our second hypothesis is that once customers were found to steal electricity, they are aware of inspections and subsequently are less likely to commit further electricity theft. Therefore, finding them or their neighbors to commit theft again is more likely in the far future and increases over time as they become less risk-aware.
As a consequence, we need to extend the Hawkes process by incorporating both, self-excitement in order to model the first hypothesis, as well as self-regulation in order to model the second hypothesis.
Only few works have been reported on modeling anomaly detection using self-excitement and self-regulation, such as faulty electrical equipment in subway systems reported in Ref. \citen{ertekin2015reactive}.

The neighborhood is essential from our point of view as neighbors are likely to share their knowledge of electricity theft as well as the outcome of inspections with their neighbors.
We therefore want to extend this model by optimizing the number of temporal processes to be used. In the most trivial case, one temporal process could be used for all customers combined. However, this would lead to a model that underfits, meaning it would not be able to distinguish among the different fraudulent behaviors. In contrast, each customer could be modeled by a dedicated temporal process. However, this would not allow to catch the relevant dynamics, as most fraudulent customers were only found to steal once. Furthermore, the computational costs of this approach would not be feasible. Therefore, we suggest to cluster customers based on their location and then to train one temporal process on the customers of each cluster. Finally, for each cluster, the conditional intensity of its temporal process at a given time can then be used as a feature for the respective customers. 
In order to find reasonable clusters, we suggest to solve an optimization problem which includes the number of clusters, i.e. the number of temporal processes to train, as well as the sum of prediction errors of all customers.

\subsection{Correction of spatial bias}
Previous inspections may have focused on certain neighborhoods. How can we reduce the covariate shift in our training set? 

The customers inspected are a sample of the overall population of customers. However, that sample may be biased, meaning it is not representative for the population of all customers. A reason for this is that previous inspections were largely focused on certain neighborhoods and were not sufficiently spread among the population. This issue has not been addressed in the literature on NTL yet. All works on NTL detection, such as Refs. \citen{nagi2010nontechnical,glauner2016large,nagi2008non,muniz2009irregularity,costa2013fraud,spiric2014using,ramos2011new,nizar2006customer,costa2013optimization} and \citen{di2012improving}, implicitly assume that the customers inspected are from the distribution of all customers.
Overall, we think that the topic of bias correction is currently not receiving the necessary attention in the field of machine learning as a whole. For about the last ten years, the paradigm followed has been labeled in Ref. \citen{banko2001scaling}: ``It's not who has the best algorithm that wins. It's who has the most data." However, we are confident to also show that having more \textit{representative} data will help rather than just having a lot of more data for NTL detection.

Bias correction has initially been addressed in the field of computational learning theory, see Ref. \citen{cortes2014domain}, which also calls this problem covariate shift, sampling bias or sample selection bias in Ref. \citen{zadrozny2004learning}. For example, one promising approach is resampling inspection data in order to be representative for the overall population of customers. This can be done by learning the hidden selection criteria of the decision whether to inspect a customer or not.  Covariate shift can be defined in mathematical terms as introduced in Ref. \citen{zadrozny2004learning}:
\begin{itemize}
\item Assume that all examples are drawn from a distribution $D$ with domain $X \times Y \times S$,
\item where $X$ is the feature space,
\item $Y$ is the label space
\item and $S$ is $\{0,1\}$.
\end{itemize}

Examples $(x, y, s)$ are drawn independently from $D$. $s=1$ denotes that an example is selected, whereas $s=0$ does not. The training is performed on a sample that comprises all examples that have $s = 1$. If $P(s\lvert x, y) = P(s\lvert x)$ holds true, we can imply that $s$ is independent of $y$ given $x$. In this case, the selected sample is biased but the bias only depends on the feature vector $x$. This bias is called covariate shift.
An unbiased distribution can be computed as follows:

\begin{align}
\label{eq:cond}
\hat{D}(x,y,s) := P(s=1)\frac{D(x,y,s)}{P(s=1\vert x)}.
\end{align}

Spatial point processes surveyed in Ref. \citen{baddeley2007spatial} build on top of Poisson processes. They allow to examine a data set of spatial locations and to conclude whether the locations are randomly distributed in a space or if they are skewed. 
Eq. (\ref{eq:cond}) requires $P(s=1\vert x) > 0$ for all possible $x$. In order to compute this non-zero probability for spatial locations $x$, we suggest to use and amend spatial point processes in order to reduce the spatial covariate shift of inspection results. This will in turn allow to train more reliable NTL predictors.

\subsection{Scalability to smart meter profiles of millions of customers}
How can we efficiently implement the models in order to scale to Big Data sets of smart meter readings?

Experiments reported in the literature range from data sets that have up to a few hundred customers in Refs. \citen{nagi2010nontechnical,sahoo2015electricity} and \citen{ramos2009fast} through data sets that have thousands of customers in Refs. \citen{cabral2009fraud} and \citen{ramos2012identification} to tens of thousands of customers in Refs. \citen{muniz2009neuro} and \citen{costa2013fraud}.
The world-wide electricity grid infrastructure is currently undergoing a transformation to smart grids, which include smart meter readings every 15 or 30 minutes. The models reported in the literature that work on smart meter data use only very short periods of up to a few days for NTL, such as in Refs. \citen{cabral2009fraud,depuru2011support,nagi2010ntl} and \citen{sahoo2015electricity}.
Future models must scale to millions of customers and billions of smart meter readings. The focus of this objective is to perform the computations efficiently in a high performance environment. For this, we suggest to redefine the  computations to be computed on GPUs, as described in Ref. \citen{abadi2016tensorflow}, or using a map-reduce architecture introduced in Ref. \citen{zaharia2010spark}.

\subsection{Creation of a publicly available real-world data set}
How can we compare different models?

The works reported in the literature describe a wide variety of different approaches for NTL detection. Most works only use one type of classifier, such as in Refs. \citen{nagi2010nontechnical,costa2013fraud,cabral2009fraud} and \citen{sahoo2015electricity}, whereas some works compare different classifiers on the same features, such as in Refs. \citen{ramos2012identification,ramos2011new} and \citen{nizar2008power}.
However, in many cases, the actual choice of classification algorithm is less important. This can also be justified by the ``no free lunch theorem" introduced in Ref. \citen{wolpert1996lack}, which states that no learning algorithm is generally better than others.

We are interested in not only comparing classification algorithms on the same features, but instead in comparing totally different NTL detection models.
We suggest to create a publicly available data set for NTL detection. Generally, the more data, the better for this data set. However, acquiring more data is costly. Therefore, a tradeoff between the amount of data and the data acquisition costs must be found. The data set must be based on real-world customer data, including meter readings and inspection results. This will allow to compare various models reported in the literature. For these reasons, it should reflect at least the following properties:
\begin{itemize}
\item Different types of customers: the most common types are residential and industrial customers. Both have very different consumption profiles. For example, the consumption of industrial customers often peaks during the weekdays, whereas residential customers consume most electricity on the weekends.
\item Number of customers and inspections: the number of customers and inspections must be in the hundreds of thousands in order to make sure that the models assessed scale to Big Data sets.
\item Spread of customers across geographical area: the customers of the data set must be spread in order to reflect different levels of prosperity as well as changes of the climate. Both factors affect electricity consumption and NTL occurrence.
\item Sufficiently long period of meter readings: due to seasonality, the data set must contain at least one year of data. More years are better to reflect changes in the consumption profile as well as to become less prone to weather anomalies.
\end{itemize}

\section{Conclusion}
\label{chapter:conclusion}
Non-technical losses (NTL) are the predominant type of losses in electricity power grids. We have reviewed their impact on economies and potential losses of revenue and profit for electricity providers.
In the literature, a vast variety of NTL detection methods employing artificial intelligence methods are reported. Expert systems and fuzzy systems are traditional detection models. Over the past years, machine learning methods have become more popular. The most commonly used methods are support vector machines and neural networks, which outperform expert systems in most settings.
These models are typically applied to features computed from customer consumption profiles such as average consumption, maximum consumption and change of consumption in addition to customer master data features such as type of customer and connection type. Sizes of data sets used in the literature have a large range from less than 100 to more than one million.
In this survey, we have also identified the six main open challenges in NTL detection: handling imbalanced classes in the training data and choosing appropriate evaluation metrics, describing features from the data, handling incorrect inspection results, correcting the covariate shift in the inspection results, building models scalable to Big Data sets and making results obtained through different methods comparable. We believe that these need to be accurately addressed in future research in order to advance in NTL detection methods. This will allow to share sound, assessable, understandable, replicable and scalable results with the research community.
In our current research we have started to address these challenges with the methodology suggested and we are planning to continue this research. We are confident that this comprehensive survey of challenges will allow other research groups to not only advance in NTL detection, but in anomaly detection as a whole.

\section*{Acknowledgments}

We would like to thank Angelo Migliosi from the University of Luxembourg and Lautaro Dolberg, Diogo Duarte and Yves Rangoni from CHOICE Technologies Holding S\`arl for participating in our fruitful discussions and for contributing many good ideas. This work has been partially funded by the Luxembourg National Research Fund.

%

\section*{References}


\bibliographystyle{unsrt}
\bibliography{references}

%
%
%
%
%

\end{multicols}
\end{document}